# Complex Wavelet – SSIM based Image Data Augmentation

Ritin Raveendran
*School of Electronics Engineering*
*Vellore Institute of Technology*
Vellore, India
ritinr96@gmail.com

Aviral Singh
*School of Electronics Engineering*
*Vellore Institute of Technology*
Vellore, India
aviralsingh17@gmail.com

Rajesh Kumar M
*School of Electronics Engineering*
*Vellore Institute of Technology*
Vellore, India
mrajeshkumar@vit.ac.in

*Abstract*— One of the biggest problems in neural learning networks is the lack of training data available to train the network. Data augmentation techniques over the past few years, have therefore been developed, aiming to increase the amount of artificial training data with the limited number of real world samples. In this paper, we look particularly at the MNIST handwritten dataset – an image dataset used for digit recognition, and the methods of data augmentation done on this data set. We then take a detailed look into one of the most popular augmentation techniques used for this data set – elastic deformation; and highlight its demerit of degradation in the quality of data, which introduces irrelevant data to the training set. To decrease this irrelevancy, we propose to use a similarity measure called Complex Wavelet Structural Similarity Index Measure (CW-SSIM) to selectively filter out the irrelevant data before we augment the data set. We compare our observations with the existing augmentation technique and find our proposed method works yields better results than the existing technique.

*Keywords—Elastic Deformation, CWSSIM, Data Augmentation, MNIST Database, Image*

## I. Introduction

Convolutional Neural Networks(CNN) are currently the state of the art machine learning technique in applications pertaining to image recognition, superseding other techniques like Convolutional Support Vector Machine(CSVM) etc. By mechanism, increasing the amount of training data increases the testing accuracy of the CNN [1]. However, manual collection of data is, especially in case of images, a tedious process, for example – collecting handwritten digits and letters for handwriting analysis.

Data Augmentation is one of the possible solutions to this problem. Data Augmentation is the technique of taking the collected training dataset and applying transformations on it to form a new set of synthetic data and appending it to the original training set.

Often during data transformation, it might be possible that the label information of the upon the transformation of a raw data has been altered, and in general these problems contain features may contain exclusive or unique features upon which the classifiers may learn. Feature-space augmentation techniques are thus used to counter this shortcoming. The Synthetic Minority Over-Sampling Technique (SMOTE) [2] is applied in feature-space as it is domain independent, finding extensive usage in medical research where a significant minority class is common in datasets. Density Based SMOTE (DBSMOTE) [3] algorithm is a modified version of SMOTE, which generates synthetic samples around the center of each class. However, due to the nature of DBSMOTE's synthetic sample generation approach, it results in overfitting of data and hence is not an ideal option for data augmentation [4].

Augmentation in data-space for image data can be applied by using transformations of existing samples while preserving label information, and validating the same information by manual observation. Elastic Deformation and affine transforms

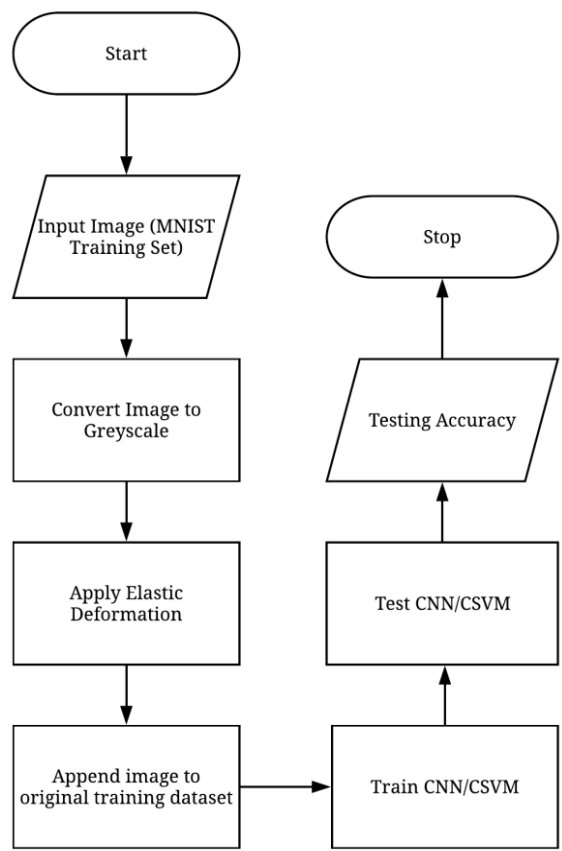

Fig.1. Generalised Training Process for CNN/CSVM

have been some of the most effective techniques under this augmentation class.

The elastic deformation [5] is a technique that transforms the data in a fashion that may produce more natural and relevant images from the existing training. A normalized random displacement field **u**(x,y) is defined that for each pixel location (x,y) in an image assigns a unit displacement vector, to ensure that $R_w = R_o + α\mathbf{u}$, where $R_w$ and $R_o$ imply the location of the pixels in the original and deformed images respectively. The intensity of the displacement in pixels is given by α. The smoothness of the displacement field is controlled by the parameter σ, which is the standard deviation of the Gaussian that is convolved with matrices of uniformly distributed random values that form the x and y dimensions of the displacement field **u**. Fig.1. shows the flowchart for the general training procedure with Elastic Deformation.

The Complex Wavelet Structural Similarity Index (CW-SSIM) is a widely accepted measure of the similarity between two images. However, its use in image classification is relatively new. It has been used in handwritten digit image classification [6]. Additionally, a KNN based CW-SSIM model has also been proposed in [7] for classifying digits from the Modified National Institute of Standards and Technology Database.

## II. ELASTIC DEFORMATION

Elastic Deformation works on the concept of applying elastic distortion to images. Through this, we obtain synthetic images which can be appended to the data set. In this section, we will look at the algorithm and an important demerit that we have found using this technique.

---

**Algorithm 1: Elastic Deformation**

**Input:** Original image
**Output:** Elastically deformed greyscale image
1. Input the original image and convert it to grey scale.
2. Create and then normalize the Gaussian field **u** of standard deviation σ.
3. Apply the scaling parameter α on the Gaussian field **u**
4. Apply the random displacement field on the greyscale image to obtain the output image

---

The effect of elastic distortions is seen in Fig. 1 where the input and output images of the algorithm are given. The random displacement map created out of the Gaussian field is given in Fig. 2.

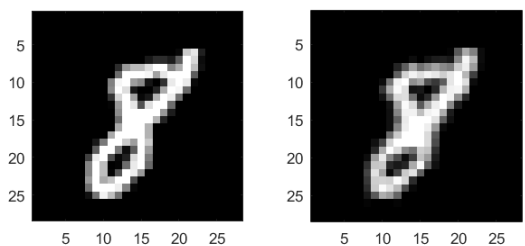

Fig. 2. Effect of Elastic Deformation (Left: Input Image, Right: Output Image)

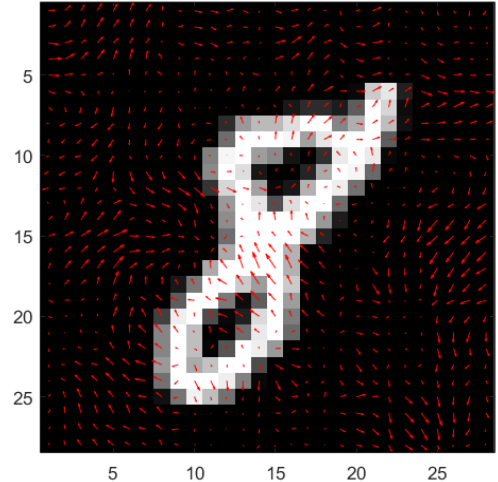

Fig. 3. Random Displacement Map on Input Image (In Red)

The demerit of using elastic deformation for handwritten digits is that in some cases, the quality of data degrades. It is seen that even for the optimum values of the parameters α and σ (found by Simard et Al. [5]) there are some distorted images which are not visually recognised to be the original image digit. Fig. 3 shows a few cases of these degraded images for a few digits.

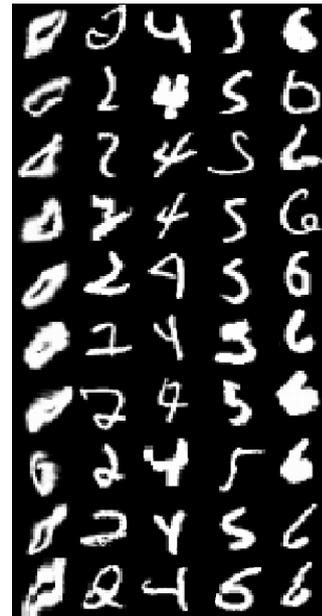

Fig. 4. Degraded Images after Elastic Distortions for 0,2,4,5,6

These degraded images from the elastic deformation technique should not be appended to the training data set as this would train the network to learn unrecognizable digits which could lower the accuracy of the network.

## III. PROPOSED ALGORITHM

In order to minimize the number of the degraded images entering our training set, we propose a new algorithm for the whole process of training. This would result in training only the best images rated in terms of digit recognizability.

---

**Algorithm 2: CW-SSIM Enhanced Training Model**

**Input:** MNIST Dataset
**Output:** Testing Accuracy of the Network
1. Input the original image from MNIST training set and convert it to grey scale.
2. Take alpha and sigma values as input for elastic deformation function
3. Apply elastic deformation on the input image for the alpha and sigma values taken in step 2.
4. Compute the CWSSIM index value between original image and the transformed image
5. If CWSSIM > Threshold **s,** then append the image to the original training dataset, else go to step 1.
6. Convert the augmented dataset into binary format
7. Initialize training of neural network with the augmented dataset as training data
8. Initialize testing of neural network with MNIST testing dataset
9. Record resulting accuracy

---

### A. Complex Wavelet Structural Similarity Index (CW-SSIM)

The CW-SSIM index is an enhancement of the Structural Similarity Index Measure (SSIM) method to the complex wavelet domain [8]. The idea here was to design a measurement that is insensitive to "non-structural" geometric distortions. The output of this measure is a value ranging from 0 (least similar) to 1(most similar).

$$CWSSIM(c_x, c_y) = \frac{2 | \sum_{i=1}^{N} c_{x,i} c_{y,i}^* | + K}{\sum_{i=1}^{N} | c_{x,i} |^2 + \sum_{i=1}^{N} | c_{y,i} |^2 + K}$$

where K is a small positive constant number taken as 0.03.

Wang et Al. states that the constant wavelet coefficient phase determines the results of CW-SSIM: "the structural information of local image features is mainly contained in the relative phase patterns of the wavelet coefficients" [8].

As elastic deformation does not deal with any large changes in translation, scaling or rotation or even blurring, we felt that the CW-SSIM Index would be the ideal filter for us in finding out which elastic images would be allowed to be a part of the training data.

## IV. EXPERIMENTAL RESULTS

The MNIST database of handwritten digits, available from [9], has a training set of 60,000 examples, and a test set of 10,000 examples. It is a subset of a larger set available from NIST. The digits have been size-normalized and centred in a fixed-size image of size 28x28. All tests are based on different scaling values of elastic deformation. Each trial was done 30 times for every value of α and the average has been recorded.

### A. Performance Evaluation of Proposed Method with CNN

The CNN is set up using a learning rate of 0.001 and 1000 epochs. Number of inputs to the network are 784 (shape of image is 28x28 = 784), and 10 output classes are corresponding to each digit from 0-9. A total of 3 convolutional layers are setup. First layer has 32 filters with a kernel size of 5 while the second layer has 64 filters with a kernel size of 3 while the third layer is the output.

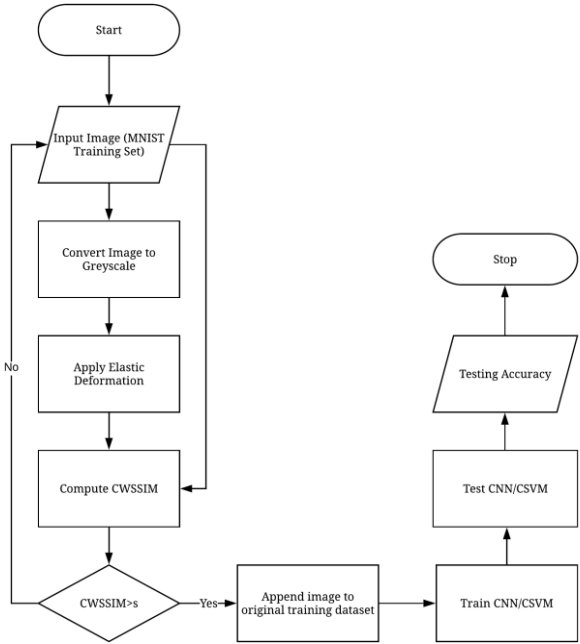

Fig. 5. Proposed Training Process

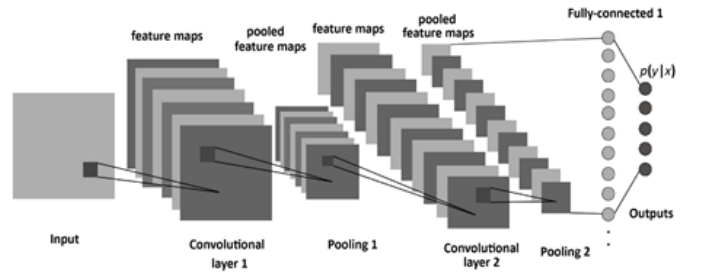

Fig. 6. Architecture of the CNN

TABLE I. ERROR RATE OF THE NETWORK VS ALPHA

| α | Error Rate (%) | | |
|---|---|---|---|
| | Existing Method(σ=34) | Proposed Method (σ=34) | |
| | Elastic Deformation (Highest) | Elastic Deformation with CWSSIM (Average) | Elastic Deformation with CWSSIM (Highest) |
| 2.0 | 0.79 | 0.76 | 0.63 |
| 2.5 | 0.83 | 0.84 | 0.79 |
| 3.0 | 0.81 | 0.83 | 0.76 |
| 3.5 | 0.95 | 0.94 | 0.82 |
| 4.0 | 0.94 | 0.89 | 0.73 |
| 4.5 | 0.98 | 0.92 | 0.86 |
| 5.0 | 0.89 | 0.89 | 0.59 |
| 5.5 | 0.94 | 0.91 | 0.6 |
| 6.0 | 0.92 | 0.9 | 0.84 |
| 6.5 | 0.85 | 0.79 | 0.66 |
| 7.0 | 0.9 | 0.8 | 0.72 |
| 7.5 | 1 | 0.84 | 0.78 |
| 8.0 | 0.88 | 0.94 | 0.68 |
| **8.5** | **0.86** | **0.69** | **0.42** |
| 9.0 | 0.98 | 0.81 | 0.77 |
| 9.5 | 0.92 | 1.02 | 0.89 |
| 10.0 | 0.91 | 0.93 | 0.71 |

### B. Performance Evaluation of Proposed Method on CSVM

SVM requires the data to be scaled first, that from [0,255] to [0,1]. A smaller size is sampled for testing. The entire data field is 70k x 784 array, each row representing pixels from 28X28=784 image. A classifier is created with Penalty parameter, C = 5 and kernel coefficient, gamma = 0.05.

TABLE II. ERROR RATE OF THE NETWORK VS ALPHA

| α | Error Rate (%) | |
|---|---|---|
| | Existing Method(σ=34) | Proposed Method (σ=34) |
| | Elastic Deformation (Average) | Elastic Deformation with CWSSIM (Average) |
| **2.0** | **9.62** | **8.54** |
| 2.5 | 10.53 | 8.56 |
| 3.0 | 11.19 | 9.33 |
| 3.5 | 10.76 | 10.28 |
| 4.0 | 12.64 | 11.61 |
| 4.5 | 15.76 | 13.29 |
| 5.0 | 14.69 | 14.67 |
| 5.5 | 15.81 | 14.98 |
| 6.0 | 16.74 | 16.76 |
| 6.5 | 17.83 | 16.91 |
| 7.0 | 15.76 | 16.02 |
| 7.5 | 18.84 | 18.82 |
| 8.0 | 19.13 | 18.99 |
| 8.5 | 18.76 | 19.03 |
| 9.0 | 20.32 | 20.86 |
| 9.5 | 23.74 | 23.81 |
| 10.0 | 25.91 | 25.66 |

## V. DISCUSSION OF RESULTS

We observe that our proposed model works better than the existing model on the CNN set. With an (α, σ) = (8.5,34) we observe an error rate of 0.42%. Even with a reduced number of training images in our training dataset, we are able to achieve a better error rate. Fig. 7. shows us some of the images eliminated using our proposed method.

However, it is seen that addition of elastic images to the CSVM results in more inaccuracy. This is because SVMs focus on preprocessing of features rather than increasing their number. For a tree based algorithm like SVM, increasing the number of features increases the possibility of correlated features which reduces the power of these learning algorithms. Even still, we observe that our model works well with respect to the existing model.

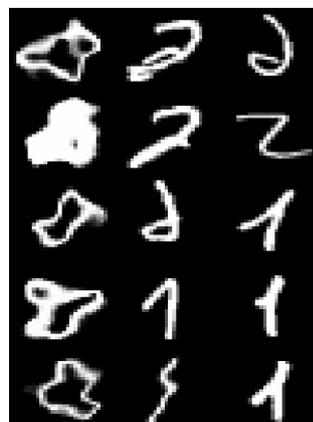

Fig. 7. Images removed by our proposed method

## VI. CONCLUSION

Elastic Deformation is one of the most popular techniques for Image Data Augmentation from the perspective of the data-space. However, this technique brings about image degradation in some cases. In this paper, we have proposed a novel model to eliminate these synthetic irrelevant images from our dataset. With this new model, the accuracy of the network for Image Classification tasks has improved. The quantitative and qualitative results obtained demonstrate the effectiveness of the proposed algorithm. In the future, the work can be focused towards extending this model onto other datasets such as the Extended MNIST or the ImageNet database.